# Not All That Is Fluent Is Factual: Investigating Hallucinations of Large Language Models in Academic Writing

Humam Khan
Department of Computer Science and Engineering, SEST
Jamia Hamdard
New Delhi, India
humamkhan.work@gmail.com

Md Tabrez Nafis
Department of Computer Science and Engineering, SEST
Jamia Hamdard
New Delhi, India
tabrez.nafis@gmail.com

Shahab Saquib Sohail
Department of Computer Science and Engineering, SEST
Jamia Hamdard
New Delhi, India
shahabsaquibsohail@gmail.com

Aqeel Khalique
Department of Computer Science and Engineering, SEST
Jamia Hamdard
New Delhi, India
aqeelkhalique@gmail.com

Rehan Hasan Khan
Energy, Resources and Industry
WSP
Calgary, Canada
rehanhasankhan009@gmail.com

*Abstract* — Large Language models (LLMs) show extraordinary abilities, but they are still prone to hallucinations, especially when we use them for generating Academic content. We have investigated four popular LLMs – ChatGPT, Grok, Gemini, and Copilot for hallucinations specifically for academic writing. We have designed 80 prompts across four categories, namely, reference generation, factual explanation, abstract generation, and writing improvement. We evaluated the model using a 0-5 rubric score, which checks factual accuracy, reference validity, coherence, style consistency, and academic tone. A novel weighted metric, Hallucination Index (HI), was introduced to measure hallucination in the responses generated by the models. Some of the most widely used evaluation metrics often fail to check errors which alter sentiment in machine-translated text. We found that Grok and Copilot perform better on reference generation tasks, but they often struggle with abstract or stylistic prompts, with HI values of 0.67 and 0.70, respectively. Whereas, Gemini and ChatGPT have done well with having stronger tone control, but they lack in writing factual tasks and higher hallucination risk with HI scores of 0.53 and 0.57, respectively. Our study found that hallucination behavior does not depend solely on model architecture but also on the type of task and the prompting conditions we are providing. We propose that our work opens new research dimensions for future researchers.

*Index Terms* — Large Language Models, Hallucination, Academic Writing, Hallucination Index

## I. INTRODUCTION

Large Language Models (LLMs) have shown advanced multimodal and language understanding capabilities. It enables strong performance in various tasks such as summarization, dialogue, image–text interpretation, and guided reasoning [1], [2]. Because of the reason that can produce coherent and context-aware responses, these models are widely used in scientific and academic writing [3]. As this is advancing with time, it has its drawbacks too. One of which is producing fabricated or hallucinated content. Hallucination continues to be a major limitation, with models producing sometimes fluent but incorrect and fabricated content [2], [4]. Hallucinations can result in factual inaccuracies in a sensitive domain such as healthcare. This can potentially affect reliability and patient safety [4]. There are other studies that suggests that hallucinated content may appear convincing and well-structured but it can increase the possibility of user misinterpretation [3].

There are several methods which have been proposed to reduce hallucinations. One of the methods includes contrastive learning to separate hallucinated and truthful representations [1]. Other study states that there are knowledge-grounding methods that use external structured knowledge for validation [2]. On the evaluation side, there's new research looking for better ways to measure how faithful these systems are [5]. They are using rubric- based scoring scales and entity-level fact assessment in summarization systems [6]. The current ways of checking AI is not very effective. They require experts to review everything manually and they don't work well for all domain too [4], [5].

These gaps help us better understand the need for checking these tools on their reliability. In areas like medicine or law, it is most important as we cannot compromise the quality on sensitive domains. Because of these major issues it is difficult to trust only on AI. To solve this, we created a clear and practical method to test how well AI performs important academic tasks, such as explaining ideas, giving correct references, and writing well-organized answers. The main aim is to understand when AI is reliable and when human verification is still necessary.

## II. RELATED WORK

Large Language models (LLMs), like ChatGPT, are advanced AI tools that can read, understand, and generate text in a very human-like way. It has been used for a lot of tasks, especially for drafting texts and generating references. Recent research shows that these models can produce hallucinated content and fabricated citations, which raises concerns about if these models are reliable or not. Safran et al. found that the references generated by ChatGPT-4 in musculoskeletal topics are only 7.5% fully accurate and 42.5% are fabricated ones. When it was verified by humans, its accuracy significantly improved, which implied that LLMs cannot be trusted alone for factual accuracy [11]. In a similar manner Pratama et al. found that targeted student training could reduce fabricated references and improve citation quality. It means that user intervention is very important, even though the study was limited in scale and domain [14].

Pattnayak et al.'s study shows that methods such as Retrieval-Augmented Generation (RAG), prompt engineering, and multimodal approaches may help reduce hallucinations, but they do not fully eliminate false or fabricated citations [12]. Edelman et al. further added to this by developing Valsci. It is a system combining RAG, chain-of-thought prompting, and bibliometric scoring to verify scientific claims at scale. Their tool significantly reduced hallucination rates, and it also increased processing speed compared to humans. It was primarily tested on benchmark claims and relies on infrastructure support [13]. Gantana et al. compared some AI writing tools, which include ChatGPT, SciSpace, and Writeless. They reported a large variation in reference reliability and content coherence. Most of the outputs contained fabricated references and weaker reasoning when compared with the human-written outputs [15]. Sharun et al. introduced a term called "artificial hallucination" to describe GPT-3.5 outputs. They found that approximately 24% of the references were fabricated, highlighting the ongoing risk in the biomedical domain [16].

Acut et al. found that prompts which are clearly specified are directly proportional to the reference accuracy. It has been noticed that prompts which are targeted produce up to 84% existing references, whereas general prompts resulted mostly with vague, incorrect and non- existent ones. Even minor differences in the prompts highly influenced the outcome [17]. Farhat et al. examined the trustworthiness of ChatGPT with the help of Bibliometric Analysis. They found that the model mainly produced inaccurate and entirely fabricated references, which further highlighted the strict need for human verification [18].

AlSagri et al. have done a comparative analysis of the scientific writing performance of ChatGPT and Gemini.They found that both models produce outputs with structure and clarity, but they vary in reference accuracy and factual grounding. It affects their suitability for academic use [19]. Sohail et al. outlined some of the key limitations of LLMs, such as hallucinations and reasoning errors. Researchers have pointed out that if LLMs are to be used responsibly in academic research, they must be supported by proper evaluation methods [20]. One of the examples we can look at is Zohery et al's description of ChatGPT. They describe ChatGPT as helpful in academic writing and for generating ideas as well [7,8]. Another study focuses more on the ethical aspect. Their focus is more on preventing plagiarism [10], even so they do not provide any proof about factually wrong information in the output which is generated. Lendvai et al's study uses scientometric analysis. It explores applications, ethical risks and even hallucinations but they do not provide directly how good the answers generated by LLMs actually are.

One study found that some common evaluation tools available such as BERTScore, BLEU, and METEOR can miss mistakes in translations. This shows that simply checking how similar two texts look is not enough. What really matters is whether the meaning and facts are correct. This problem is similar to what we see with LLMs, they often produce convincing answers that may still contain wrong information [22]. There are many studies which depends upon automatic evaluation. Even though many studies depend on it, we cannot fully trust them. There are some cases in which even evaluation which was done without re-checking scored high even though the text was incorrect [21]. This makes it clear that we need better ways to detect hallucinations.

These studies reflect that even though LLMs can make academic writing faster, they still are producing hallucinated content. There are some strategies to double check the content to reduce errors but they cannot be fully trusted. This gap leads to an important research question: how can we create a practical evaluation framework that can detect hallucinations in AI-generated academic writing across different categories and different types of models?

## III. METHODOLOGY

### *A. Models*

The methodology was designed in a manner that systematically measure and compare hallucination behavior between four widely used Large Language Models (LLMs): Grok, Copilot, Gemini, and ChatGPT. The goal was to check how each of the given models perform in academic writing. Especially the tasks which are prone to hallucinations such as reference generation, factual explanation, and abstract writing. To maintain fairness and reproducibility, all models were evaluated under the same conditions. We set the temperature to 0.2 to limit randomness in the output produced by the LLMs, also, all responses were limited to 100 words. No model was finetuned or modified; each was evaluated in its default public configuration to reflect real-world usage.

### *B. Prompt Benchmark*

We designed 80 prompts across four categories:

1. Reference Generation (fabrication-prone)
2. Factual Explanation (logic and correctness)

3. Abstract Generation (high creativity; high hallucination risk)
4. Writing Improvement (tone preservation)

Prompts included both real and trap-based items designed to induce fabricated evidence. A benchmark of 80 prompts was created and categorized into four types. The Reference Generation category focused on citation-related tasks, where models often produce inaccurate references in the responses. Factual Explanation prompts judged the logical basis and correctness of the models' knowledge. Abstract Generation prompts required a creative blend, a known source of hallucination in generative systems due to their blend of creativity and factual expectations. Writing Improvement prompts tested the models' ability to polish text while preserving original meaning and tone. The benchmark intentionally included both realistic academic tasks and "trap-based" prompts designed to provoke hallucinations by inserting misleading claims or unverifiable assumptions. These four prompt types were prepared under two broader categories based on the type of hallucination trigger: (1) Hallucination Bait Prompts, which target fabrication tendencies, and (2) Sycophancy Trap Prompts, which test whether models agree with misleading or incorrect user assumptions. A hybrid approach was used to collect the data, with outputs from the other models being manually gathered while Gemini responses were generated using an API key.

### *C. Scoring Rubric*

Using a scoring rubric intended to identify various types of hallucinations, we assessed each model's output. We used five criteria to check the responses namely, Coherence (C), Style Consistency (S), Reference Validity (R), Factual Accuracy (F_d), and Academic Tone (W). We checked these by scoring them from 0-5. A score of 0 meant there were serious problems, and 5 meant the answer was strong and it met academic expectations.

While going through the responses, we paid close attention to accuracy and checked if the references actually existed and if they were formatted properly. We also considered how easy the response was to follow and whether it kept the same academic style throughout. To make the scoring fair, three different reviewers evaluated each response on their own. We then averaged their scores. The model names were hidden, and we followed clear guidelines to avoid bias as much as possible for the results we produced.

### *D. Hallucination Index*

To measure hallucination, we created our own weighted Hallucination Index (HI) instead of relying only on existing evaluation metrics. Since academic writing depends heavily on correct facts and real references, we gave the highest weight to factual accuracy (4) and reference validity (3). We gave coherence a medium weight (2) because the response still needs to be clear. Style consistency and academic tone were given the lowest weight (1 each), as they are less important compared to factual correctness in research writing. These weights were chosen based on what matters most in academic work. In this, accuracy and proper citation are more critical than writing style.

We added all the weights, 4, 3, 2, 1, and 1, which gave us 11. This total was used in the denominator so that the final score would be properly balanced. After calculating the weighted average, we converted it into a 0–100 scale by multiplying it by 20. Doing this makes the results easier to read and compare across different models. If a model gets a higher Hallucination Index (HI), it means it makes fewer mistakes and performs better overall. The HI gives a clear and practical way to judge how reliable a model is, especially for academic writing where accuracy matter the most.

$$HI = \left( \frac{4F_d + 3R + 2C + S + W}{11} \right) \times 20$$

When we evaluated the AI, we checked five main things. The most important was factual accuracy, we made sure the facts were correct, because wrong information makes the whole answer useless, no matter how well it is written. Reference validity was checked since fake sources reduce trust. Coherence was another key factor, we looked if the answer was clear and easy to follow. We also looked at style consistency, checking whether the writing maintained a regular academic style. Finally, we checked if the Academic tone sounded formal and proper for academic work.

## IV. EXPERIMENTAL RESULTS

*a) Overall Performance of the models:* Figure 1 shows clear differences in Hallucination Index (HI) scores among the four models. Copilot achieved the highest score (70.74%), meaning it produced fewer hallucinated responses in our evaluation. Next is Grok with 67.86% and showed fairly steady behavior across tasks. ChatGPT scored 57.04%. Its answers usually sounded clear , but it sometimes gave incorrect references, which lowered its score. Gemini calculated score was the lowest , 53.73%, mainly because it made more mistakes in citations. AI tools don't make the same number of mistakes. Some do a better job, while others make more errors. It really depends on how each AI creates its answers and how carefully it handles the information it gives.

Table 1 : Mean Scores per Model

| Model | F_d | R | C | S | W | HI (%) |
|---|---|---|---|---|---|---|
| Grok | 3.64 | 4.25 | 2.43 | 2.47 | 3.03 | 67.86 |
| Copilot | 3.60 | 3.81 | 3.72 | 3.28 | 2.48 | 70.74 |
| Gemini | 3.00 | 2.25 | 2.55 | 3.10 | 2.60 | 53.73 |
| ChatGPT | 3.21 | 1.52 | 3.81 | 3.15 | 3.47 | 57.04 |

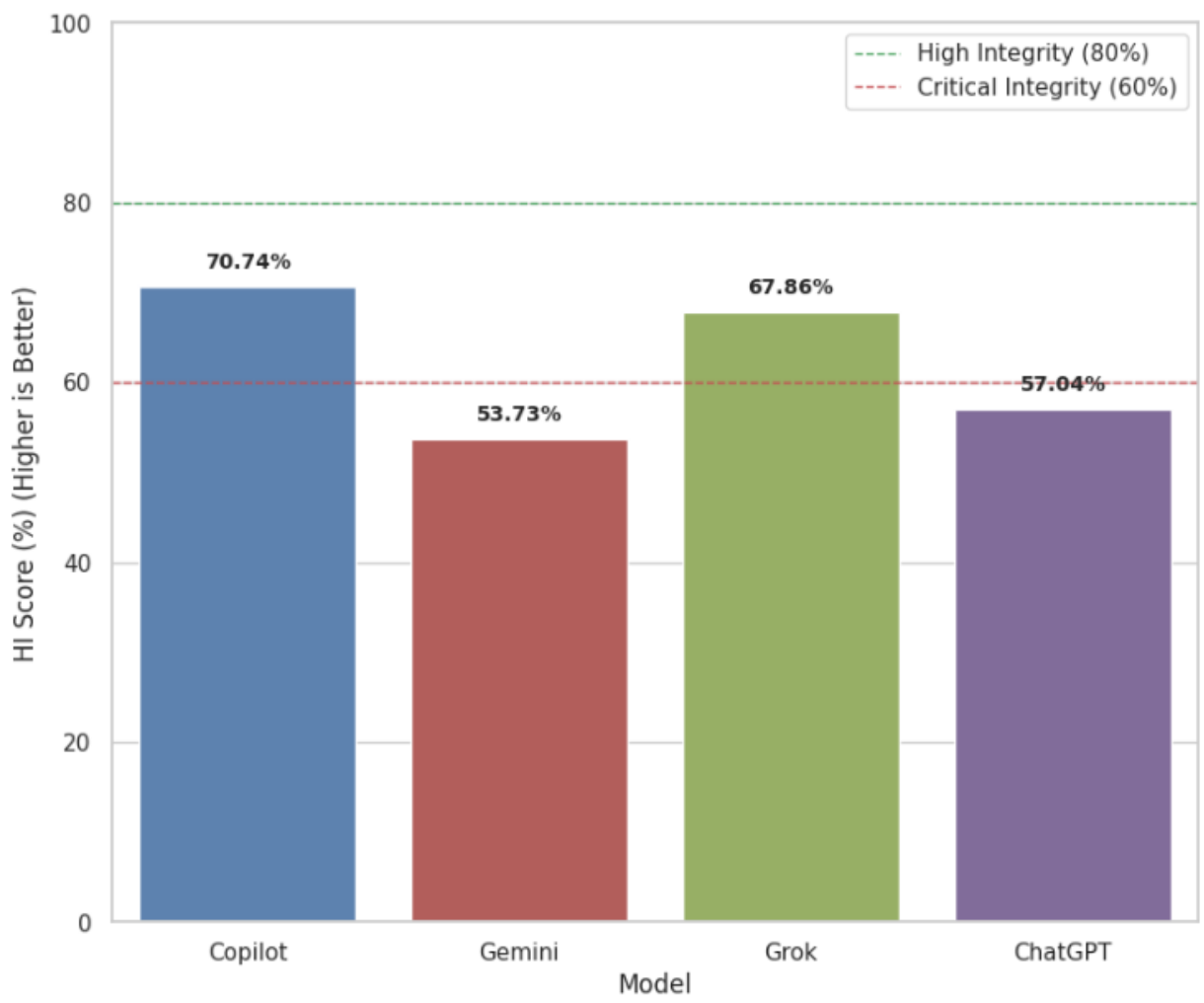

*Figure 1: Overall HI per Model*

b) *Performance Across 5 Metrics* - Figure 2 shows that each model's performance across the five different categories (F_d, R, C, S, W). Here, Copilot scores high in coherence, style, and reference validity. Grok results show that it is strong for factual accuracy and that we can rely on references. But it is moderate for style consistency. ChatGPT performs better in style (S) and academic tone (W). It shows that it is better in fluent generation. But it performs weaker in reference validity (R). Gemini shows that it is the least reliable in Factual Accuracy and Reference Validity. This comparison highlights that hallucinations increases not only from factual errors but also from citation weaknesses. These are the areas where Gemini and ChatGPT perform poorly as compared to Copilot and Grok.

Table 2: Detailed Mean Scores per Model and Category

| Model | Reference Generation | Factual Explanation | Abstract Generation | Writing Improvement |
|---|---|---|---|---|
| | (F_d/R / HI%) | (F_d / R / HI%) | (F_d / R / HI%) | (F_d / R / HI%) |
| **Grok** | 4.4 / 4.5 / 78.0 | 4.1 / 3.9 / 66.09 | 2.4 / 4.4 / 61.73 | 4.9 / 4.6 / 77.09 |
| **Copilot** | 4.3 / 4.2 / 77.5 | 4.7 / 5.0 / 88.37 | 3.5 / 2.8 / 60.73 | 1.8 / 3.3 / 57.73 |
| **Gemini** | 3.9 / 3.7 / 74.0 | 2.8 / 2.1 / 55.18 | 1.4 / 2.7 / 41.27 | 3.5 / 2.0 / 54.54 |
| **ChatGPT** | 4.0 / 3.6 / 75.0 | 5.0 / 1.1 / 64.00 | 1.6 / 2.0 / 54.91 | 3.5 / 1.0 / 54.00 |
| **Average** | **4.2 / 4.1 / 76.5** | **4.0 / 3.5 / 68.2** | **2.5 / 2.8 / 55.4** | **3.1 / 2.9 / 62.1** |

c) *Task-Wise Comparison of HI* - Figure 3 shows values of HI per model across four different task categories.

- Reference Generation: Grok (HI=78.0) and Copilot (77.5) perform the best. It shows that it has a very strong citation grounding.
- Factual Explanation: Here, Copilot scores the highest HI of 88.37. It further shows that it works best for its reasoning ability.
- Abstract Generation: All models score lowest here (avg. HI≈55%), which confirms that creative tasks induce more hallucinations as it includes more false content.
- Writing Improvement: Grok scores the highest HI (77.09), which indicates that it is best for writing.

The numerical data in Table 2 support these observations. It shows that Grok and Copilot scores good in tasks which requires high precision. Gemini and ChatGPT performs poorly when working with reference-dependent prompts.

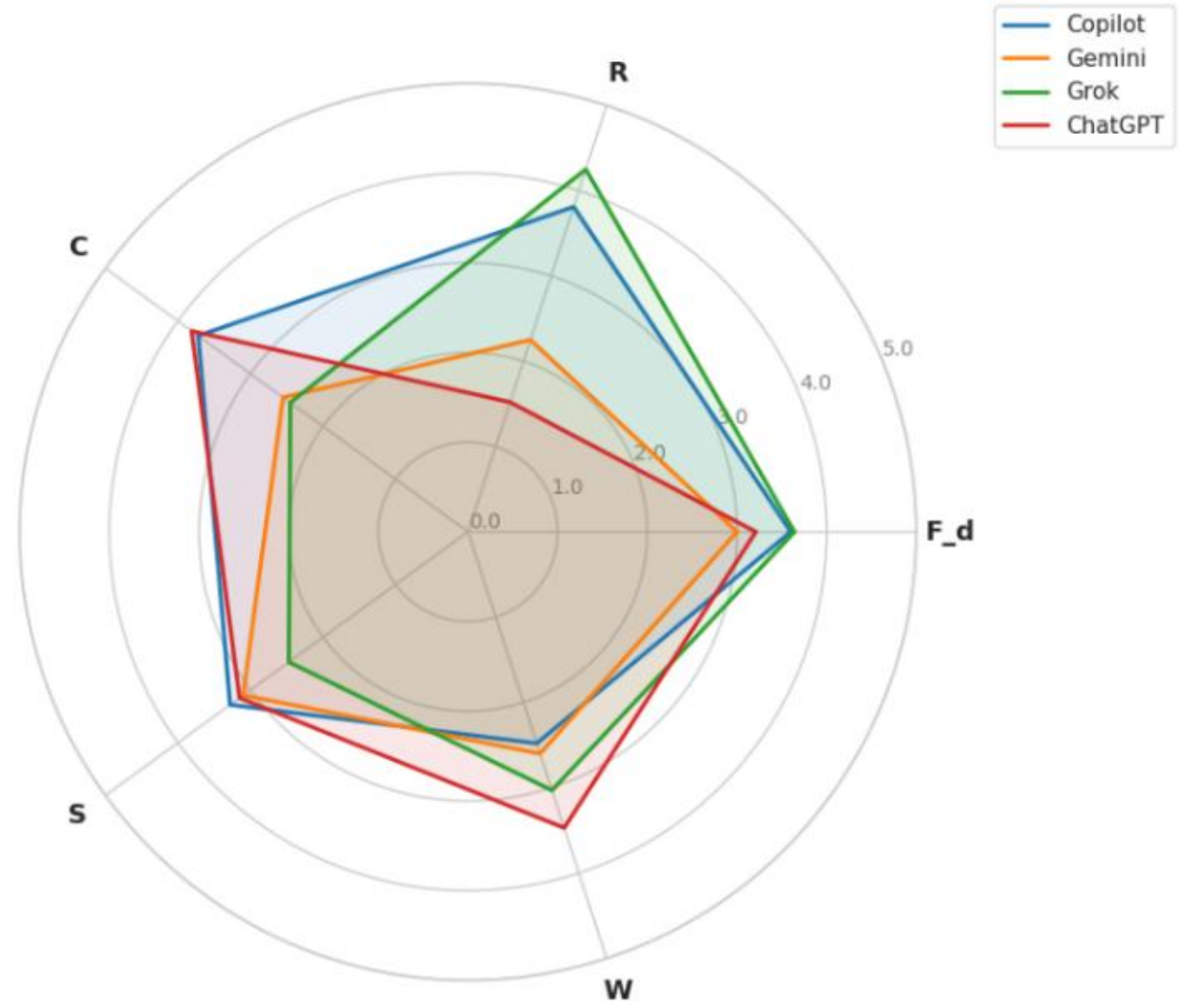

*Figure 2: Performance Profile Radar (5 Metrics)*

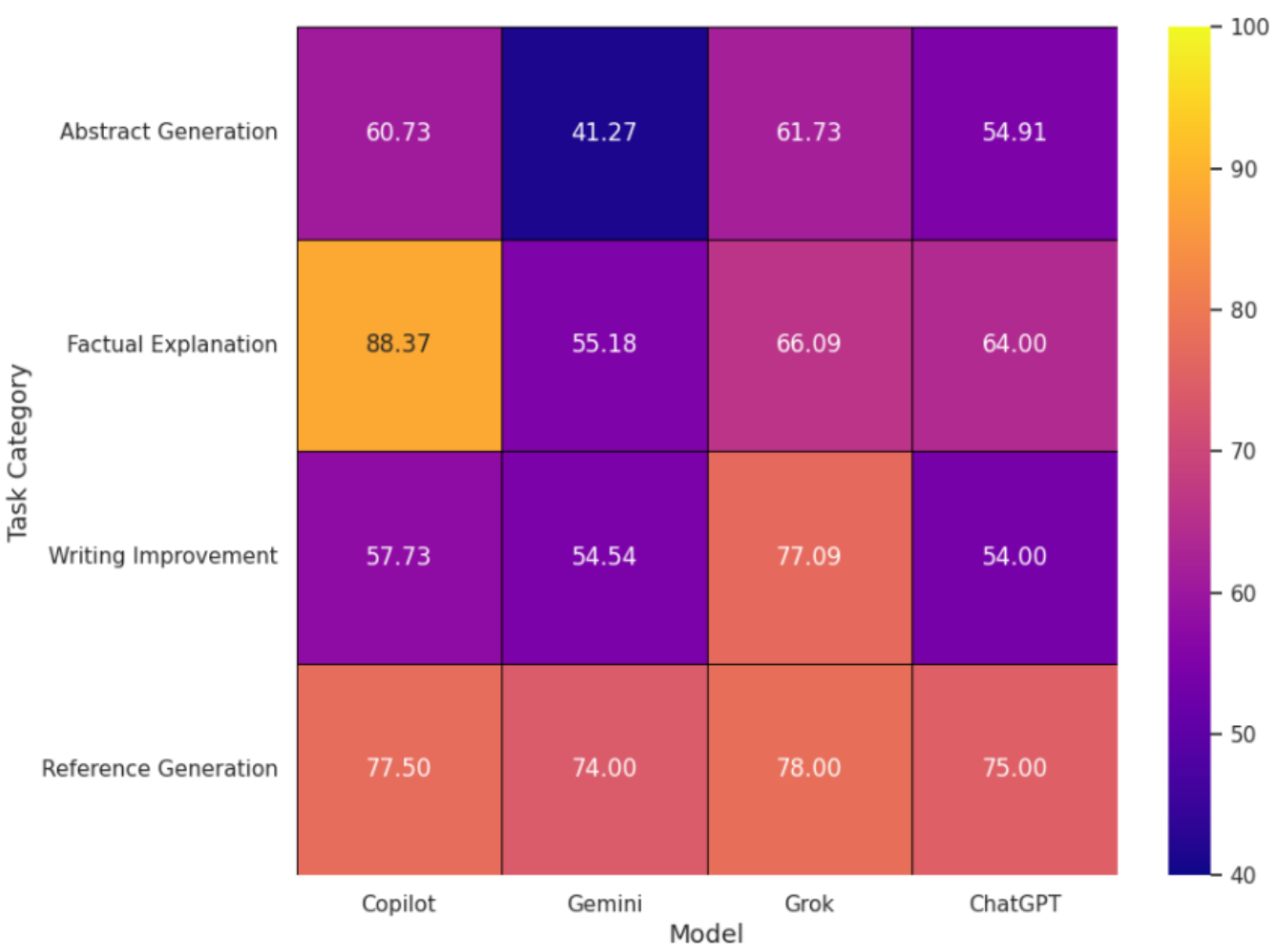

*Figure 3: Heatmap - HI (%) by Task Category*

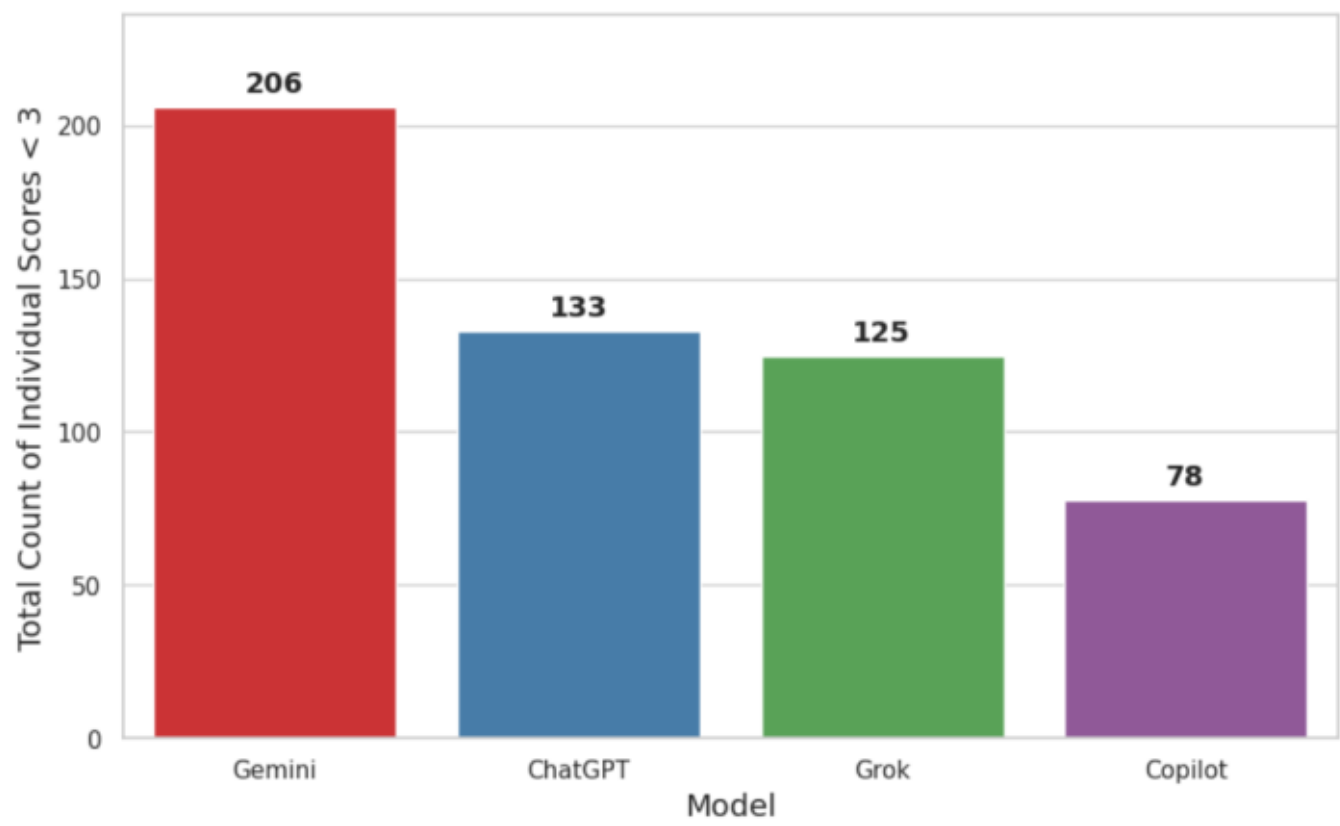

*Figure 4: Frequency of Low-Quality Scores (Score<3) per Model*

d) *Frequency of Low-Quality Scores* - Figure 4 shows the number of low-quality evaluations (scores <3). Gemini shows the highest count (206 occurrences), which indicates frequent hallucinations. ChatGPT (133) and Grok (125) fall in the midrange, which shows that they are moderately stable. We see problems mostly in factual accuracy. Here, Copilot shows the lowest count (78 occurrences). It further explains why it is most reliable in diverse tasks instead of focusing on just one task. We found that the fewer the low-quality responses, the more academically dependable the model is.

## V. DISCUSSION

The results reveal that hallucinations vary significantly across models and task types. Grok and Copilot show superior integrity in citation-dependent and reasoning-intensive tasks, likely due to better internal checking or retrieval-style mechanisms. Gemini and ChatGPT, while strong in fluency and style, rely heavily on internal parametric memory, which increases the likelihood of generating fabricated references or unsupported claims. The Hallucination Index successfully captured these disparities, demonstrating strong alignment with human evaluator judgments. Tasks involving creativity or unverifiable assumptions, such as abstract writing, produce the highest hallucination frequencies, highlighting the need for external fact-checking when using LLMs in scholarly workflows.

## VI. CONCLUSION

This study provides a structured evaluation of hallucination tendencies across four major LLMs using an 80-prompt benchmark and a weighted hallucination scoring framework. The findings demonstrate that Copilot and Grok offer stronger reliability for academic writing, particularly in factual explanation and reference-based tasks. Gemini and ChatGPT remain useful for stylistically rich writing but require additional verification of factual content and citations. The results reinforce that LLMs can assist academic work effectively but should not replace manual validation in scholarly settings.

Future research can expand this evaluation by incorporating multilingual prompts and exploring automated hallucination scoring using model-as-judge or self-consistency techniques. Additionally, domain-specific weighting schemes, particularly for biomedical, legal, and engineering contexts, may provide more specialized insight. Continuous monitoring of updated LLM versions will also be essential, as rapid improvements in architecture and grounding mechanisms may significantly shift hallucination patterns over time.

## ACKNOWLEDGMENT

Disclosure of Delegation to Generative AI

The authors declare the use of generative AI in the research and writing process. According to the GAIDeT taxonomy (2025), the following tasks were delegated to GAI tools under full human supervision:

- Proofreading and editing

The GAI tool used was: ChatGPT- 4.5.

Responsibility for the final manuscript lies entirely with the authors.

GAI tools are not listed as authors and do not bear responsibility for the final outcomes.

Declaration submitted by: 'Collective Responsibility'